\documentclass[cameraready]{Interspeech}
\title{Looking for Affect in Spontaneous Finnish Speech through Linguistic Interpretability}

\author[affiliation={1}]{Kalle}{Lahtinen}
\author[affiliation={2}]{Liisa}{Mustanoja}
\author[affiliation={1}]{Okko}{Räsänen}
\address{
    $^1$ Signal Processing Research Centre, Tampere University, Finland \\
    $^2$ Research Centre Plural, Tampere University, Finland  
}

\email{kalle.t.lahtinen@tuni.fi, liisa.mustanoja@tuni.fi, okko.rasanen@tuni.fi}

\keywords{speech emotion recognition, affect in language, computational paralinguistics}

\newcommand{\rev}[1]{\textcolor{black}{#1}}
\usepackage{comment}
\usepackage{float}
\usepackage{ulem}
\begin{document}

\maketitle

\begin{abstract}
    % Tällä hetkellä 993
    Existing research on affect in speech has shown how acoustic surface characteristics and content-related linguistic aspects of speech both relate to perceived emotional arousal and valence. However, it is not clear what the relative contributions of these two factors are in the perceptual process. This is especially true for Finnish, for which most existing studies focus on either acoustic-phonetic or text analysis. This paper presents a study where we systematically explore the combinatory role of text- and audio-based features in modeling the human perception of valence and arousal using a newly released affective speech corpus for spontaneous Finnish. We show that the combination of text- and audio-based features improves valence regression results over the individual modalities, whereas for arousal regression the complementary effect is not substantial. The results support prior findings from other languages, providing new data and knowledge on spontaneous Finnish speech.
    %Existing research on affect in speech has shown how acoustic-phonetic and linguistic aspects of the input both relate to perceived emotional arousal and valence. However, it is not clear to what extent the perception is driven by acoustical versus content-based linguistic properties of the input. This is especially true for Finnish, for which most existing studies focus on either acoustic-phonetic or text analysis. This paper presents a study where we systematically explore the combinatory role of text- and audio-based features in modeling the human perception of valence and arousal using a newly released affective speech corpus for spontaneous Finnish. We show that the combination of text- and audio-based features improves valence regression results over the individual modalities, whereas for arousal regression the complementary effect is not substantial. The results support prior findings from other languages, providing new data and knowledge on spontaneous Finnish speech.
\end{abstract}

\section{Introduction}

Emotional valence and arousal \cite{russell1980circumplex, russell1989cross} are commonly used annotation targets for studying affect in both spoken \cite{scherer2003vocal, larrouy2025sound} and written \cite{aff_indonesian,EilolaTiinaM.2010Anf2,soderholm2013valence, linden2023finnsentiment} language. While acted speech data is frequently paired with discrete emotion labels (e.g., happy, angry, sad) in the field of speech emotion recognition (SER) (see, e.g., \cite{rathi2024analyzing}), continuous valence and arousal scores are often chosen for spontaneous (unacted, \textit{in-the-wild}) speech \cite{Lotfian, lahtinen2025finnaffect}. This is since continuous ratings are considered to enable the study of more nuanced differences in affect in language as opposed to clear-cut discrete emotion classes \cite{naini25_interspeech}. The use and perception of affective markers in speech are also culture- and language-dependent \cite{scherer2003vocal, waaramaa2013perception}. Even though the complementary role of linguistics and paralinguistics related to linguistic affect is well established (see, e.g., \cite{wagner2023dawn}), it is not clear how much of the variation of perceived valence and arousal in spontaneous speech can be explained with the different modalities of speech in a given language, nor is it conclusive what the complementary interactions of the different feature types are. 

In this paper, we examine to what extent implicit and explicit text- and audio-based features carry information related to the perception of valence and arousal in spontaneous Finnish speech. We use a recently published spontaneous speech corpus, FinnAffect \cite{lahtinen2025finnaffect}, in our analysis, containing speech audio samples with matching text transcriptions and with ratings of valence and arousal from human listeners. We analyze the relative contributions of various audio- and text-based feature sets and their combinations in predicting human ratings of valence and arousal. As the main finding, we show that valence perception in the context of spontaneous Finnish is strongly related to the complementarity of text- and audio-based features, whereas for arousal, the acoustical signal carries most of the meaningful information.

\section{Prior Work and Motivation}

Previous studies have shown that affective information related to speaker arousal is mainly carried and perceived through the acoustic properties of the speech signal (\textit{how the speaker speaks}), whereas valence has been considered to be more related to the linguistic semantic content of the expressed message (\textit{what the speaker says}) \cite{wagner2023dawn,calvo2010affect}. Furthermore, it has been shown that purely audio-based SER models capable of state-of-the-art valence prediction implicitly learn linguistic semantic structures from the acoustical training data \cite{wagner2023dawn}. 

Until recent works in \cite{vaaras2023development, lahtinen2025finnaffect}, earlier studies of acoustic-phonetic aspects of affect in spoken Finnish have largely focused on acted data, with a relatively small number of speakers and a narrow linguistic scope in the spoken material used in the analysis (see, e.g., \cite{laukkanen1996physical,toivanen2003automatic,airas2006emotions,waaramaa2010perception}). Larger data has mainly been applied in text-based research on sentiment analysis for the FinnSentiment dataset \cite{linden2023finnsentiment}, subsequent  development of a sentiment analysis tool \cite{finnsentiment_huggingface}, and in recent speech emotion analyses based on speech transcriptions from the Parliament of Finland \cite{tarkka-etal-2024-automated, ristila2026hopesfearsemotion}, but with little relation to the acoustics of the speech.

In linguistic studies, the affective roles of syntactic \cite{Visapaa2013, Ahonen2019} and lexical \cite{EilolaTiinaM.2010Anf2, soderholm2013valence, ohman2022selffeil} properties of Finnish have been examined to some extent, but within specific or constrained contexts. The collection of so-called emotion lexicons (lists of words paired with affect-related human annotations) is a common practice for approaching affect in many languages (e.g., \cite{aff_indonesian}), but it is uncertain how well lexical analysis alone can be utilized in the affective analysis of real-life speech. While looking at the function of independent infinitives as affective and empathetic constructs in Finnish, Visapää \cite{Visapaa2013} states that not all expressions describing emotional situations can be considered affective, as language provides many means for non-affective expression for communicating emotional content. She argues that the concept of linguistic affect becomes relevant in interactive contexts, where the language is used to express emotions and attitudes about what is being said without explicitly stating any emotional states.

The advancement of scientific understanding of linguistic affect in Finnish speech calls for further analysis based on larger speech corpora, enabling the examination of different communicative contexts and interactions while considering both the acoustic-phonetic features and verbal contents (going beyond simple lexeme- or syntax-based structures) of speech concurrently.

\section{Methods}

We conducted a feature importance experiment by testing how different combinations of text- and audio-based features affect the performance of a multilayer perceptron (MLP)-based regression model for predicting human listener ratings of perceived valence or arousal in Finnish utterances. In the pool of possible features, we used both implicit (latent embedding vectors from pretrained neural network models) and explicit (feature vectors with interpretable dimensions) features\rev{, with 127 feature combinations in total}. We also analyzed the relationship of the interpretable feature dimensions in relation to the perceived valence scores of the data samples.

In addition to using text and audio samples from the dataset for feature extraction, we also included the human-perceived arousal annotation score as one possible feature dimension when training the valence model. This is since prior work has suggested that the perception of arousal and valence may interact with each other \cite{waaramaa2010perception, goudbeek2010beyond}, and hence grounding of valence perception in concurrent arousal may help the models in fitting to human data. Perceived valence was used similarly when training the model for arousal prediction.

\subsection{Dataset}
\label{sec:dataset}

We used the annotated portion of the recently published FinnAffect speech corpus as our dataset \cite{lahtinen2025finnaffect}. The annotated section of the corpus contains 12,000 speech samples of 1--20 seconds each, paired with matching text transcriptions. Each sample is annotated for continuous valence and arousal in the range of [-1.0, 1.0] by five human annotators (listeners). Out of the annotated samples, 2,000 were annotated by all five listeners (which we call the Gold Standard (GS) subset), and the remaining 10,000 samples were annotated by individual listeners in separate sets of 2,000 samples for each annotator. The corpus is derived from three existing Finnish source corpora, which all contain spontaneous (dialogue or topically prompted monologue) speech. The source corpora used are the \textit{Lahjoita Puhetta} (LP) ("donate speech") \cite{LP}, the Longitudinal Corpus of Finnish Spoken in Helsinki (\textit{Helpuhe}, HP) \cite{helpuhe1_en} and the Longitudinal Corpus of Finnish Spoken in Tampere (\textit{Tampuhe}, TP). 

LP is a crowdsourced speech corpus containing donated speech from over 20,000 individual speakers with 1,687 hours of transcribed speech. In the recordings (lasting from a few seconds to several minutes), the subjects speak freely about various topics prompted by the data collection software. The subjects' own devices were used for the recording. The corpus contains speech from a diverse group of people, covering different genders, age brackets, and socioeconomic backgrounds. The HP and TP datasets contain interview recordings from the 1970s, 1990s, and 2010s, which have been collected for the purpose of conducting sociolinguistic longitudinal studies on the variation of spoken language in the regions of Helsinki and Tampere. The original interview recordings are approximately one hour long each, containing speech from over 200 individuals.

For the experiments, we use the same principles for splitting the data into training, validation, and test sets as used in the regression tests in \cite{lahtinen2025finnaffect}. The data split properties are described in Table \ref{table:dataset}. Division of samples into testing and training was based on the speaker IDs of the annotated samples. The entire GS set was used as the testing dataset, whereas the training and validation sets contain samples from a distinct set of speakers, with each sample having one human annotation. In the samples originating from the HP and TP corpora, the speaker ID information is based on the original recording filenames, resulting in a different ID for the same speaker recorded during a different decade or in several session recordings from the same decade. Taking this into account, the number of unique speakers was 733 and 3095 in the GS and Train+Val sets, respectively.

\begin{table}[t]
\centering
%\vspace{-6pt}
\caption{Dataset split statistics. All data refers to the annotated portion of FinnAffect, GS is the data subset used for model testing, and Train+Val refers to the combined training and validation subset. Gender U refers to unknown speaker gender.}
%\vspace{-9pt}
\includegraphics[width=0.47\textwidth]{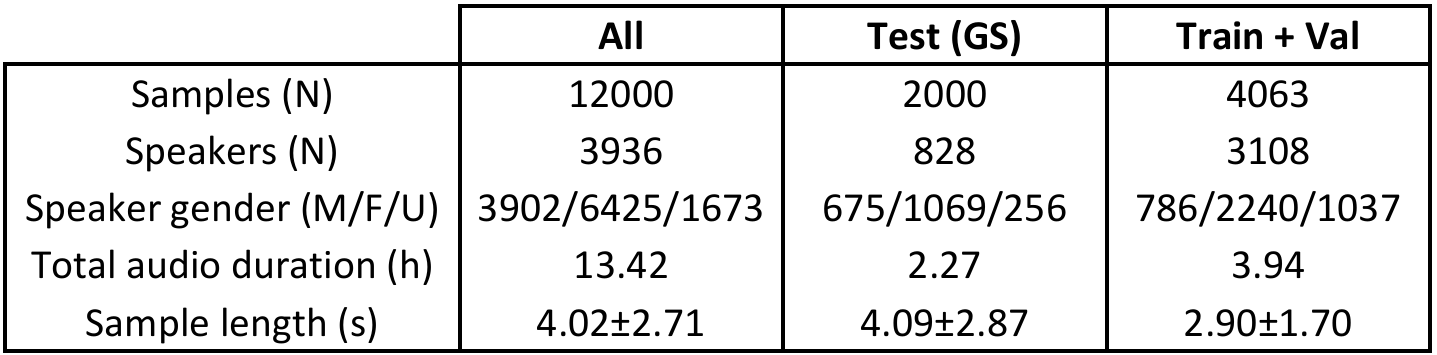}
\label{table:dataset}
\vspace{-25pt}
\end{table}

\begin{table*}[t]
\centering
\vspace{-10pt}
\caption{Valence regression results \rev{(CCC mean $\pm 1$ standard deviation)}. The top three feature combinations based on train+val performance are reported along with the individual features and combinations of all text and all audio features. CF - SF is the average performance gain for using colloquial instead of standardized text transcripts (n.s. stands for non-significant differences). %\rev{significant (\( \dag  \) marks p $< 0.10$, \( \ddag  \) marks p $< 0.05$)}.  
CF and SF text samples are marked with ° and *, respectively. Footnote T marks text-based and A audio-based features. Ann. SD refers to test set CCC variability across individual annotators.}
%\vspace{-9pt}
\includegraphics[width=\textwidth]{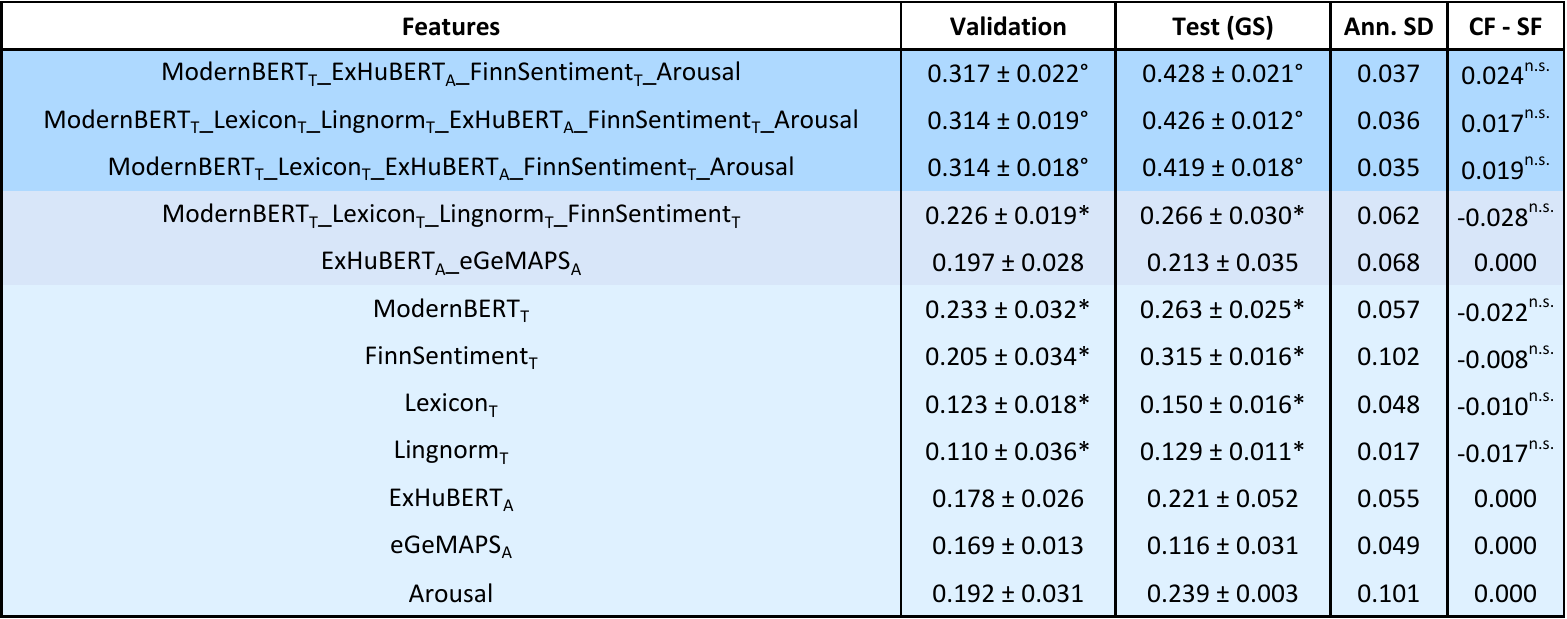}
\label{table:valence_results}
\vspace{-25pt}
\end{table*}

%However, the overlapping speaker ID counts are minimal in relation to the total number of speakers (less than 5\% of individual speakers for the GS set and less than 1\% for the training and validation set) and hence considered inconsequential for the present analyses.

\subsection{Data processing and studied feature sets}
\label{sec:feature_extraction}

Both text- and audio-based features were extracted from the annotated samples. The original text data were transcriptions of the recordings with spoken content transcribed as it was said (i.e., the transcriptions were colloquial language instead of standard Finnish). For all text-based features, we used both colloquial Finnish ("CF") transcriptions and transcriptions that were standardized to standard Finnish ("SF") using the GPT 4.1 API \cite{openaigpt4.1}. We ran all the experiments with both transcription styles separately. The text data was preprocessed with the Finnish Trankit pipeline \cite{nguyen2021trankit}, from which the token-level lemmatization, morphological, and syntactic features were utilized when constructing explicit text-based feature vectors. In addition, the continuous arousal annotation score was used as an optional one-dimensional feature when training the model for valence regression, and vice versa, simulating a scenario where the listener is making concurrent evaluations on both dimensions.

From the text data we extracted the following features for all utterances: i) mean-pooled 1024-dim token embeddings encoded by the pretrained Finnish ModernBERT model \cite{reunamo2025pretrainingfinnishmodernberts} (ModernBERT), ii) sentiment analysis softmax posteriors (negative, neutral, positive) from the FinBERT-FinnSentiment model \cite{finnsentiment_huggingface} (FinnSentiment), iii) a concatenated emotional lexicon-based 36-dim feature vector based on four available lexicons for Finnish  (Sentiment and Emotion Lexicon for Finnish (SELF), Finnish Emotion Intensity Lexicon (FEIL) \cite{ohman2022selffeil}, valence and arousal ratings for 420 Finnish nouns by age and gender \cite{soderholm2013valence} and affective norms for 210 British English and Finnish nouns \cite{EilolaTiinaM.2010Anf2}) (Lexicon), and iv) a normalized 106-dim (CF) or 107-dim (SF) linguistic feature vector (Lingnorm) based on the Trankit pipeline. 

The Lexicon feature vector was computed as follows. For every lemma in an utterance, we first checked for matches in each of the lexicons. For all the matching lemmas, we computed a mean vector over the matching feature dimensions defined in the lexicon (for some lemmas, there could be several matches for one lemma in one lexicon). We then summed the lemma-specific vectors for each utterance and used the mean scores of the lemma vectors as the feature vector for the whole utterance. Additionally, we concatenated the number of lemmas that had a match in the lexicon to the utterance feature vector. From the SELF lexicon, we used the positive, negative, anger, anticipation, disgust, fear, joy, sadness, surprise, and trust dimensions; from the FEIL lexicon, we used the emotion intensity score, anger, anticipation, disgust, fear, joy, sadness, and trust dimensions; and from the 210 Finnish nouns lexicon, we used the mean and standard deviation (SD) of familiarity, mean and SD of valence, mean and SD of emotional charge, mean and SD of offensiveness, and mean and SD of concreteness dimensions. From the 420 Finnish nouns lexicon, we used the mean and SD of all valence scores and the mean and SD of all arousal score dimensions.

For the Lingnorm feature vector, we used the mean of one-hot-encoded linguistic features across the tokens of each utterance. The included features included all the possible values for the Case, Clitic, Connegative, Derivation, PronType, InfForm, NumType, Number, PartForm, Reflex, Voice, VerbForm, Tense, Polarity, Degree, Mood, Person, upos, and xpos features. We also computed the depth of the universal dependencies dependency tree for each utterance based on the parsing by Trankit. 

From the audio data, we extracted the mean pooled 1024-dim embeddings from the pretrained speech emotion recognition model ExHuBERT, for which Finnish was not included in its training data \cite{exhubert}. We did not fine-tune ExHuBERT for Finnish, as we wanted to ensure that the representations focused only on non-Finnish, acoustic aspects of affect expression. We also extracted 88-dim eGeMAPSv02 functionals using the OpenSmile toolkit \cite{opensmile} to represent low-level acoustic-phonetic characteristics of the speech that have been earlier identified as relevant for affective speech analysis \cite{egemaps}. The individual dimensions in the eGeMAPS feature vector were z-score normalized across all the data. 

\begin{table*}[t]
\centering
\vspace{-10pt}
\caption{Arousal regression results \rev{(CCC mean $\pm 1$ standard deviation)}. The top three feature combinations based on train+val performance are reported along with the individual features and combinations of all text and all audio features. CF - SF is the average performance gain for using colloquial instead of standardized text transcripts \rev{\ (\dag \ marks significant difference at p $< 0.05$; paired t-test)}.  CF and SF text samples are marked with ° and *, respectively. Footnote T marks text-based and A audio-based features. Ann. SD refers to test set CCC variability across individual annotators.}
%\vspace{-9pt}
\includegraphics[width=\textwidth]{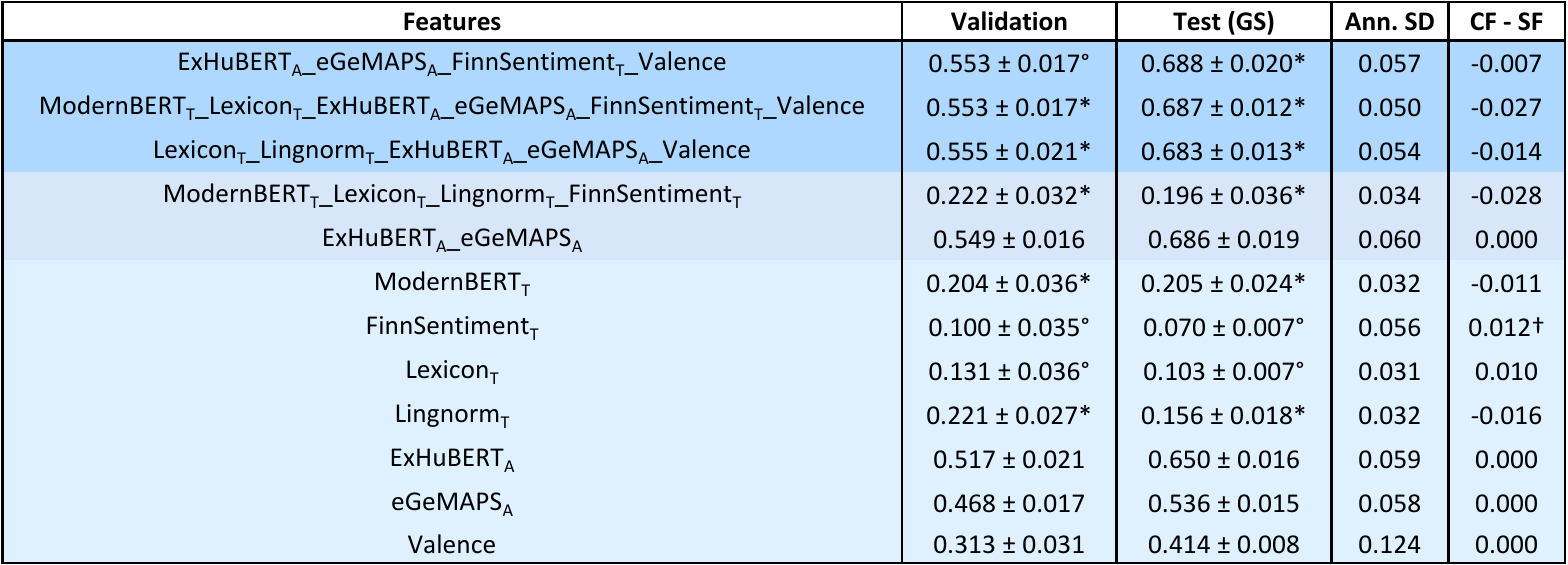}
\label{table:arousal_results}
%\vspace{-12pt}
\vspace{-20pt}
\end{table*}

\subsection{Regression setup for feature analysis}

\begin{figure}[h!]
\centering
\includegraphics[width=0.47\textwidth]{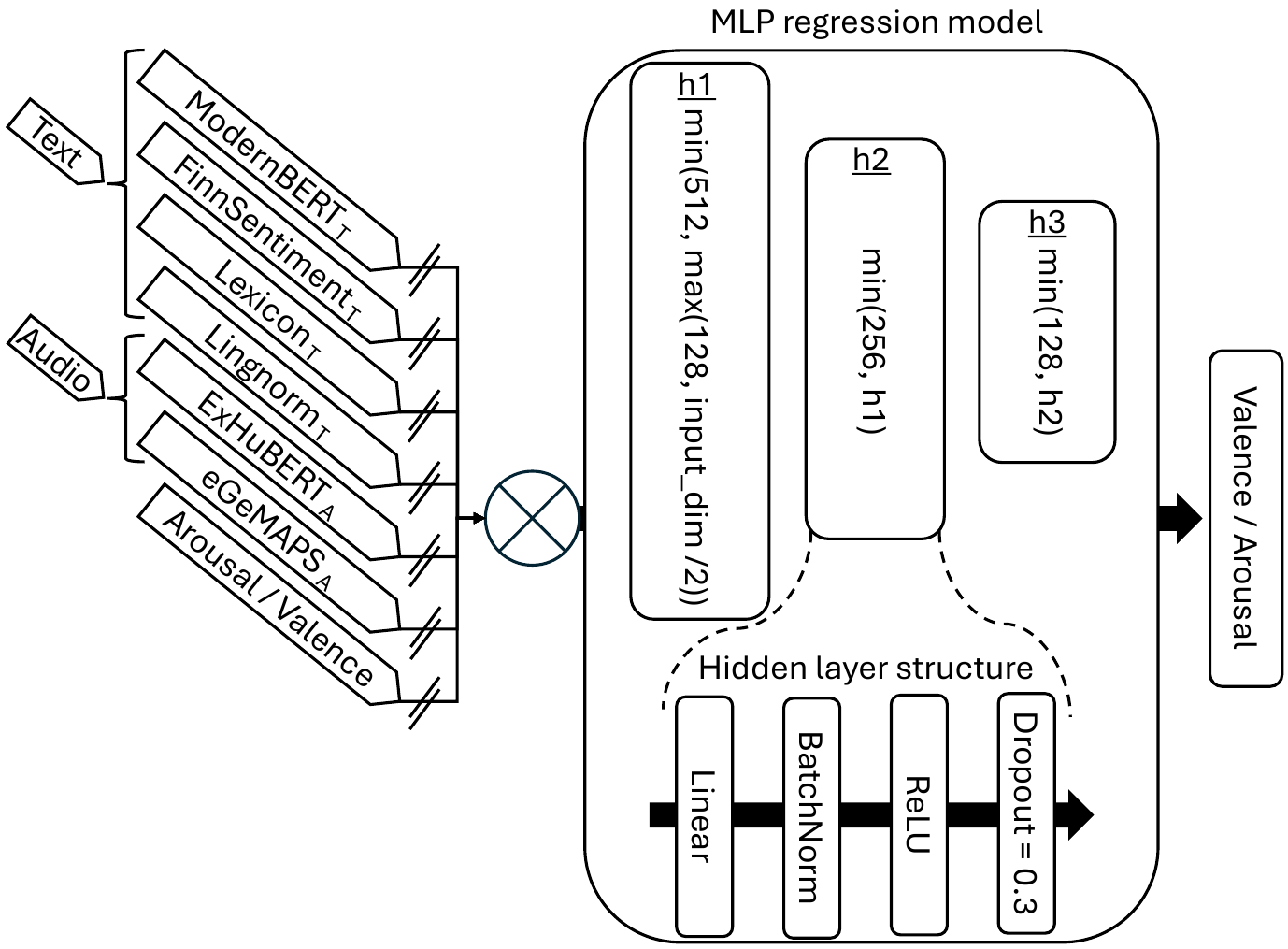}

\caption{Regression experiment setup for predicting continuous valence and arousal scores, including the model hidden layers (h1 -- h3) and the included feature sets.}
\label{fig:MLP_regression}
\vspace{-15pt}
\end{figure}

To test the relative and combined roles of different features, we used our training and validation data to train a pool of MLP regression models for both valence and arousal using all possible combinations of the features (including the individual feature sets). The structure of the experiments and the architecture of the model are described in Figure \ref{fig:MLP_regression}. We used the PyTorch library \cite{pytorch} for model implementation. 

The concordance correlation coefficient (CCC) was used as the loss function ($\mathcal{L} = 1 - CCC $) between the predicted and actual affect ratings and as the evaluation metric. Adam was used as the optimizer with a learning rate of $10^{-3}$. The training and validation were done by group folding the samples into five folds using the speaker ID as a grouping criterion (GroupKFold method in \cite{sklearn}). In all cases the model was trained for 50 epochs, and the best-performing model (in validation) was used for final testing with the GS dataset. The validation and testing were done for all the folds, and we report the mean and standard deviation ($\pm$) of CCC scores of each feature combination across the folds for both validation and testing.

For the sake of space, we report results only for the \rev{three best} performing feature combinations after  ranking the results according to the mean validation performances for all the experiments using CF and SF text data. In addition, we separately report performance for the combination of all text-based features, for all audio-based features, and for each individual feature to illustrate the total information available in each modality and in each feature type. We also report the standard deviation of annotator-specific CCCs across the annotators to illustrate the variability in listener use of different feature types \rev{and the difference in performance between CF and SF text data features.}
%We also report the standard deviation of the \rev{mean} test scores \rev{when using} individual \rev{listener annotations} in the test set to illustrate the variability in listener (annotator) use of different feature types. 
\rev{A link to the full set of results is available in Acknowledgements.} 

\section{Results}

%\subsection{Regression}

The results of the valence regression \rev{tests} are presented in Table \ref{table:valence_results}. The three best-performing models combining both audio and text features achieved test scores of CCC $>0.4$, the best feature combination of ModernBERT and ExHuBERT embeddings together with the FinnSentiment posteriors and the arousal annotation score achieving a CCC of $0.428\pm0.021$ on the test data. The text-only and audio-only CCC results were $0.266\pm0.030$ and $0.213\pm0.035$, respectively. This shows that the combination of text-based and audio-based features improves the valence regression results substantially when compared to text- or audio-based regression alone. The top three results were achieved with colloquial transcripts, \rev{although the differences between the CF and SF were not statistically significant CCCs across the 5 test folds ($p > 0.05$; paired t-test)}. However, the performance with text-only \rev{feature combinations} was generally better by CCC of $0.023\pm0.019$ \rev{(p $< 0.05$)} with SF transcripts in comparison to CF when compared across all text feature sets. %\sout{across all the tested feature combinations}. 

Results for arousal prediction \rev{tests} are presented in Table \ref{table:arousal_results}. The three best-performing models all achieved a test CCC $>0.68$. The best feature combination (ExHuBERT embeddings with the eGeMAPS features, FinnSentiment posteriors, and valence ratings) achieved a CCC test result of $0.688\pm0.020$. The text-only and audio-only CCC results were $0.196\pm0.036$ and $0.686\pm0.019$, respectively. Notably, even with the ExHuBERT embeddings alone, the model achieved a CCC test result of $0.650\pm0.016$. In this case, the top result was achieved with standardized text data, although the CF-SF difference is only significant for FinnSentiment features (p $< 0.05$).  %However, the differences between \rev{CF} and \rev{SF} data tests \rev{were again non-significant except for FinnSentiment}. 

For both arousal and valence regression, the inclusion of the \textit{other dimension} of the arousal-valence space in the training data improved the regression CCC performance by $0.029\pm0.036$ (CF) and $0.030\pm0.029$ (SF) for valence \rev{(p $< 0.01$ for CF and SF)} and by $0.025\pm0.065$ (CF) and $0.028\pm0.066$ (SF) for arousal \rev{(p $< 0.01$)}, highlighting the partial correlation of the two dimensions in the data (see also \cite{waaramaa2010perception, goudbeek2010beyond}). 

Analysis of the CCC scores across individual listeners showed that the largest subjective differences in terms of valence were for FinnSentiment and arousal features (CCC SDs of 0.102 and 0.101 across the listeners, respectively). In contrast, across-annotator variability clearly reduced for valence (SD $< 0.038$) for the top three feature combinations. For arousal, the extent that valence alone predicted arousal ratings varied substantially more across the listeners than any of the other features. The top features generally varied more across the listeners in comparison to most of the individual features. 

Finally, we also sought further interpretability of the features by performing linear regression and decision tree experiments with the Lexicon and Lingnorm features. However, given that these feature sets had only limited explanatory power over the valence and arousal ratings (Tables \ref{table:valence_results} and \ref{table:arousal_results}), it turned out to be difficult to draw any conclusions regarding the syntactic or lexical properties associated with affective content. More annotated data would be needed to systematically study the interactions of individual features with basic statistical analyses.

%The regression performance when using the interpretable text-based features (namely the Lexicon and Lingnorm feature vectors) was low. For both features individually, the CCC performance in the regression experiments was $\leq 0.150$ and $\leq 0.163$ for valence and arousal, respectively. This could be explained by the sparsity of information within the computed features. Overall, of 12,000 annotated samples, 81.1\% and 79.8\% of cells were zero-valued in the lexicon and lingnorm feature matrices. However, both feature types appear in the top three performing feature combinations in both valence and arousal experiments. 

\section{Discussion and Conclusions}

%In this study, we examined the relationship of text-based semantic and linguistic features, audio-based acoustic speech features, and the perceived emotional valence and arousal in spontaneous Finnish speech. We conducted a feature importance experiment with a set of different implicit and explicit speech features computed from both speech transcriptions and audio samples and tested how different feature combinations perform when training an MLP model for valence and arousal regression.

The current results indicate that the verbal content of Finnish spontaneous speech carries complementary information to the acoustic content in the perception of the valence of the spoken expression. 
In an earlier study by  \cite{lahtinen2025finnaffect} with the same data and data split, the authors reported CCC scores of $0.270$ for valence and $0.689$ for arousal using audio-based regression models from the 2024 MSP-Podcast Emotion Challenge \cite{MSPChallenge} after fine-tuning them for the present FinnAffect training data. In the current experiments, the valence regression results improved substantially (CCC = $0.428 \pm 0.021$) when using both audio and text-based features of the speech. In contrast, the arousal regression performance did not improve  (CCC = $0.688 \pm 0.020$) in comparison to the results in \cite{lahtinen2025finnaffect}. These findings show that text-based information is more crucial for valence perception than it is for arousal, the latter being mainly driven by the acoustic-phonetic properties of the signal. 

Our results support the earlier premise on the role of spoken content in the interpretation of valence in speech \cite{wagner2023dawn}, strengthening the relevance of text-based linguistic features in explaining variance in perceived valence in Finnish. More studies are required to get more information on the key linguistic factors related to affect in spontaneous Finnish and on valence perception in particular.

\section{Acknowledgments}
\label{sec:acknowledgements}
\ifcameraready
     The work of KL was funded by the CONVERGENCE project, a grant awarded by the Jane and Aatos Erkko Foundation to Tampere University. The authors would also like to thank Tampere Center for Scientific Computing for the computational resources used in this study. The experimentation and analysis source codes, as well as computed features, annotation scores, and analysis results are available at \url{https://github.com/SPEECHCOG/LookingForAffect/}
\else
    The experimentation and analysis codes, as well as computed features, annotation scores, and results, will be published.
\fi

\section{Generative AI Use Disclosure}

Generative AI was not used in the writing of this paper. Generative AI was used in a minor role to support programming of the experiments (syntax support, debugging, and code verification). The authors are fully responsible for all the work and content reported in the paper.

\bibliographystyle{IEEEtran}
\bibliography{mybib}

\end{document}